\pdfoutput=1

\documentclass[11pt]{article}

\usepackage{acl}

\usepackage{times}
\usepackage{latexsym}

\usepackage{microtype}
\usepackage{graphicx} 
\usepackage{float} 
\usepackage{subfigure} 
\usepackage{wrapfig}
\usepackage{graphicx}
\usepackage{subfigure}
\usepackage{amsfonts}
\usepackage{bm}
\usepackage{array,amsmath}
\usepackage{amssymb}
\usepackage{dsfont}
\usepackage{float}
\usepackage{graphicx}
\usepackage{algorithm}
\usepackage{algorithmicx}
\usepackage{algpseudocode}
\usepackage{pgf,tikz}
\usepackage{mathrsfs}
\usepackage{multirow}
\usepackage{booktabs}
\usetikzlibrary{arrows}
\usepackage{threeparttable}
\usepackage[inline]{enumitem}

\def\figref#1{Figure~\ref{fig:#1}}
\def\figlabel#1{\label{fig:#1}\label{p:#1}}

\def\tabref#1{Table~\ref{tab:#1}}

\def\tablabel#1{\label{tab:#1}\label{p:#1}}

\def\secref#1{\S\ref{sec:#1}}
\def\seclabel#1{\label{sec:#1}}

\newcounter{notecounter}

\newcommand{\enoteson}{\long\gdef\enote##1##2{{
			\stepcounter{notecounter}
			{\large\bf \hspace{1cm}\arabic{notecounter} $<<<$ ##1: ##2 $>>>$\hspace{1cm}}}}}
\enoteson
\usepackage{xcolor}
\usepackage{soul}
\usepackage{xspace}
\newcommand{\hlc}[2][yellow]{{%
		\colorlet{foo}{#1}%
		\sethlcolor{foo}\hl{#2}}%
}
\usepackage{color}

\def\fz{Frozen\xspace}
\def\pf{PromptFuse\xspace}
\def\bp{BlindPrompt\xspace}

\long\def\eat#1{}

\usepackage{microtype}

\title{Modular and Parameter-Efficient Multimodal Fusion with Prompting}
\author{Sheng Liang, Mengjie Zhao, Hinrich Sch\"utze\\
	Center for Information and Language Processing (CIS) \\
	LMU Munich, Germany\\
	{\tt \{shengliang,mzhao\}@cis.lmu.de}}

\begin{document}
\maketitle

\begin{abstract}
Recent research has made impressive progress 
in large-scale multimodal pre-training.
In the context of the rapid growth of model size, 
it is necessary to seek efficient and flexible methods other than finetuning.
In this paper, we propose to use prompt vectors to align the modalities.
Our method achieves comparable performance to several
other multimodal fusion methods in low-resource settings.
We further 
show that our method is modular and parameter-efficient for processing tasks involving two or more data modalities.
\end{abstract}

\section{Introduction}

The success of large-scale
pretrained language models
(PLMs; \citet{devlin-etal-2019-bert,xlnet,gpt3,t5paper}) 
and
image encoders
\citep{dosovitskiy2021an,liu2021swin}
has stimulated a surge of
pretrained \emph{multimodal models} 
\citep{Lu2019ViLBERTPT,tan-bansal-2019-lxmert,radford2021learning,lin2021m6} 
that align text with data in other modalities.

The fast-growing number of parameters in
the pretrained models
encourages
researchers to create more data- and parameter-efficient methods
than finetuning \citep{houlsby2019parameter,zhao-etal-2020-masking,bitfit,prefixtuning,he2022towards}.
Recently, prompting 
-- concatenating manually designed
prompt phrases \citep{schick2020s,tam2021improving,le2021many,zhao-schutze-2021-discrete}
or trained embedding vectors \citep{prefixtuning,lester2021power}
to the text input of PLMs
-- has become an important research direction.

Following this trend, \citet{Tsimpoukelli2021MultimodalFL}
introduce \emph{\fz}, 
successfully
extending PLMs into
few-shot learners (i.e., models that perform
well with only a handful of data)
for multimodal tasks, 
by pretraining a vision encoder
whose outputs are prompts fed to the PLM.
\fz performs strongly on
low-resource visual question answering
through GPT3-style  \citep{gpt3} priming (in-context learning).
\fz consists of two components:
A vision encoder (VE) (in their case, NF-ResNet-50 \citep{brock2021high})
and
an off-the-shelf PLM like GPT3.
When pretraining \fz, the
PLM takes 
the image representations
extracted by VE as prompts,
to generate captions describing the input image.
PLM parameters are \emph{fixed} and
VE is pretrained from scratch.
The success of \fz shows the  potential 
of prompting-based systems for
solving multimodal
tasks \citep{zhou2021learning,yang2021empirical,salaberria2021image}.

	\begin{figure}[t]
		\centering
		\includegraphics[width=\linewidth]{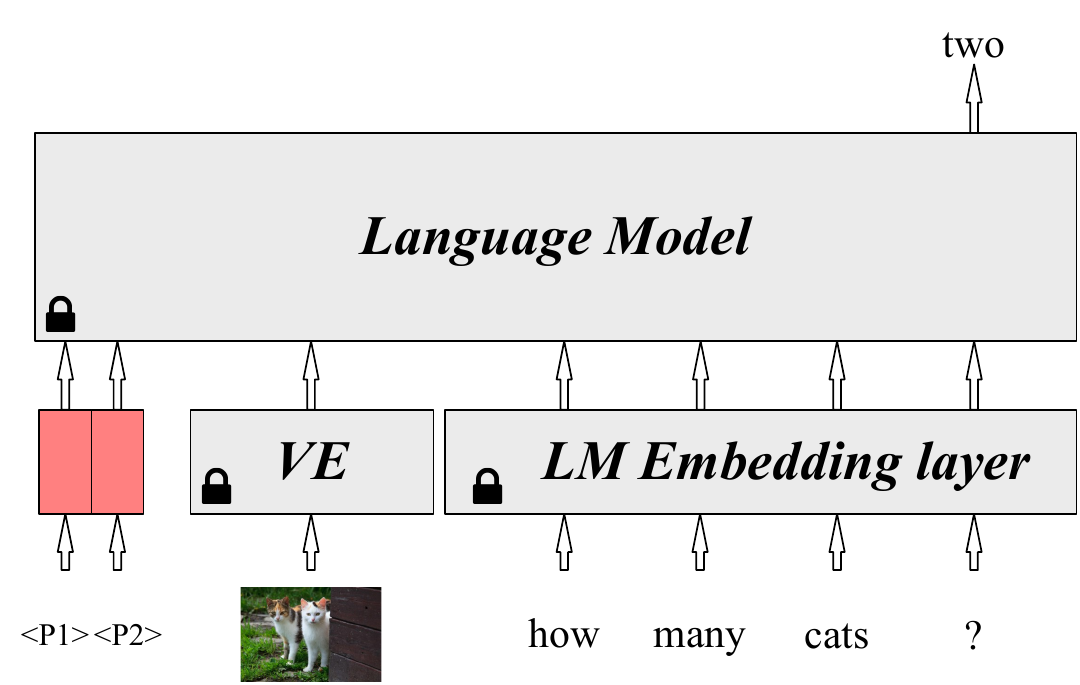}
		\caption{
			Model architecture.
			We disentangle VE's functionality by introducing
			prompt vectors.
			The only work of VE is to extract image representations.
			PLM and VE are fixed (grey) during training;
			two prompt vectors are
			the only trainable parameters (red).
		}
		\figlabel{model}
	\end{figure}

	One inherent discrepancy between \fz and prompting for NLP
	tasks \citep{prefixtuning,lester2021power} is that the
	prompt vectors
	in \fz represent part of the input, the image:
	They are image features
	extracted by VE.
	In contrast, prompt vectors in NLP
	are agnostic to the input texts:
	They are trainable parameters of the PLM embedding layer
	to be optimized during training.
	Recall that the PLM in \fz is fixed when pretraining VE.
	This implies that
	VE's trainable parameters serve
	two quite distinct purposes:
	(i) extract high quality image representations;
	(ii) align the image and text representation spaces.

	We investigate the efficacy
	of \emph{disentangling} the functionality of VE.
	Concretely, we fix
	the parameters of PLM and VE, and
	allocate extra free parameters
	for learning the
	alignment between spaces of different
	modalities when conducting a multimodal task;
	this is achieved by introducing additional
	prompt vectors.
	As a result, VE can
	dedicate itself to extract high quality
	image representations.
	We hypothesize that
	disentanglement has two benefits.
	First,
	\emph{higher modularity}
	is achieved
compared to \fz
because VE is freed from
	the objective of aligning modalities.
	Higher modularity brings higher flexibility, which is not applicable in systems like \fz:
	We can easily change the type of VE, e.g.,
	replacing a CNN with a Transformer;
	adding extra modalities like speech
	data is made possible as well.
Our architecture meets the desideratum stated by
	\citet{srivastava2014multimodal}:
	It should be possible to modularly add
	modalities to an existing multimodal system.
	Second, higher \emph{parameter efficiency}
	is achieved by fixing the
	encoders of different modalities
during training;
	the prompt vectors are the
	only module to be trained
	for aligning the representation spaces.
		
	We present \textbf{\pf},
	a prompting-based approach
	extending PLMs to multimodal tasks in a modular and efficient manner.
	Our contributions:
	(i) We show that
the prompting paradigm 
of utilizing PLMs \citep{liu2021pre}
	effectively strengthens PLMs with the ability of processing
	data in modalities besides text.
	With only $\approx$15K trainable parameters,
	\pf performs comparably
	to several multimodal fusion methods
	in low-resource regimes.
	(ii) We further propose \textbf{\bp}, which
	enforces that the
	prompt vectors  solely focus on task-specific
	information and is therefore
less prone to overfitting.

	\section{Related Work}
	\textbf{Prompting}
        is a more data- and
	parameter-efficient method of using
	pretrained language models
	(PLMs; \citet{devlin-etal-2019-bert,xlnet,gpt3,t5paper})
	than finetuning \citep{devlin-etal-2019-bert}.
	Concretely,
	\citet{gpt3}, \citet{schick2020s},
	\citet{tam2021improving}, \citet{le2021many},
        and \citet{gao2020making}
	show that prompting  outperforms finetuning
	in many NLP tasks when  annotations are
	limited, i.e., in \emph{few-shot learning}.
	\citet{prefixtuning} introduce
	prefix-tuning, only updating the
	prompt vectors, keeping the PLM fixed.
	\citet{lester2021power} introduce
	prompt-tuning -- a
	simple form of prefix-tuning --
	achieving performance
    comparable to finetuning when scaling up the number of parameters in PLMs.
	As large PLMs remain unchanged
	during prefix- and prompt-tuning,
	high parameter-efficiency is achieved.

	\textbf{Multimodal pretraining}.
	The success of
PLMs and pretrained image encoders
\citep{dosovitskiy2021an,liu2021swin}
encourage fast developments of
multimodal pretraining, 
e.g., large-scale neural networks that 
	align texts with data in other modalities
	like image \citep{
		tan-bansal-2019-lxmert,su2019vl,Cho2021UnifyingVT,
		wang2021simvlm,kim2021vilt},
	video \citep{sun2019videobert}
	and speech \citep{bapna2021slam}.
	
	Prompting methods for multimodal models were recently devised.
	\citet{zhou2021learning} learn continuous prompt vectors rather than
	natural language descriptions to model visual concepts.
	\citet{yao2021cpt} mark image regions as prompts,
	adapting pretrained vision-language models to downstream tasks.
In \fz, for a fixed PLM,
	\citet{Tsimpoukelli2021MultimodalFL} 
 pretrain a VE with image captioning 
	where image representations from the VE are used as prompt vectors.
	The VE in \fz needs to achieve two objectives:
	Extracting high quality image representations
	and properly aligning image/text spaces.
	In this work,
	we show that
	disentangling the two functionalities
	-- instead of pretraining a VE like \fz,
	we utilize pretrained VE as feature extractor
	and train
	prompt vectors to fuse the modalities --
	results in
	a more modular and efficient
	multimodal system.

	\section{Prompting as Multimodal Fusing}

	We propose to decompose the functionality of VE in \fz
	into:
	(i) providing high quality image representations to the PLM;
	(ii) aligning the image and text spaces for a multimodal task.
	Achieving (i) is straightforward --
	we leverage off-the-shelf pretrained image encoders, e.g.,
	Vision Transformer (ViT; \citet{dosovitskiy2021an}).
	We align the two representation spaces
	by 
	prompt-tuning \citep{prefixtuning,lester2021power},
        i.e.,
        by introducing prompt vectors.
	Concretely, we randomly initialize $N$ trainable vectors in
        the embedding layer of PLM.
	When processing downstream multimodal tasks,
	we \emph{finetune the prompt vectors
	but fix PLM and VE}.
	\figref{model} illustrates our model.
	We call our method 
	\textbf{\pf}.
	Having very few
	trainable parameters,
	\pf is well suited for
	low-resource regimes.

	We design a special
	attention mask for the PLM encoder,  shown in \figref{mask}.
	While the attention of input data remains fully visible,
	we enforce prompt vectors to only access each other 
	but be blind to the input data.
	We refer to this variant of
	\pf as \textbf{BlindPrompt}.
        \bp fuses 
	data in all modalities 
	using the prompt vectors
	in self-attention layers.	
        This  further
	emphasizes that
	prompt vectors
	should be focusing on
	the \emph{alignment} between
	modalities rather than on \emph{specifics} of the content
	of a modality. As a result, \bp is more
	robust to spurious statistical cues \citep{niven-kao-2019-probing}.
	For example, given a picture that dogs run after a man,  
	overfitting systems tend to answer ``poodles'' in response to the question ``What do dogs chase?''.

	\begin{figure}[t]
		\centering
                \includegraphics[width=.8\linewidth]{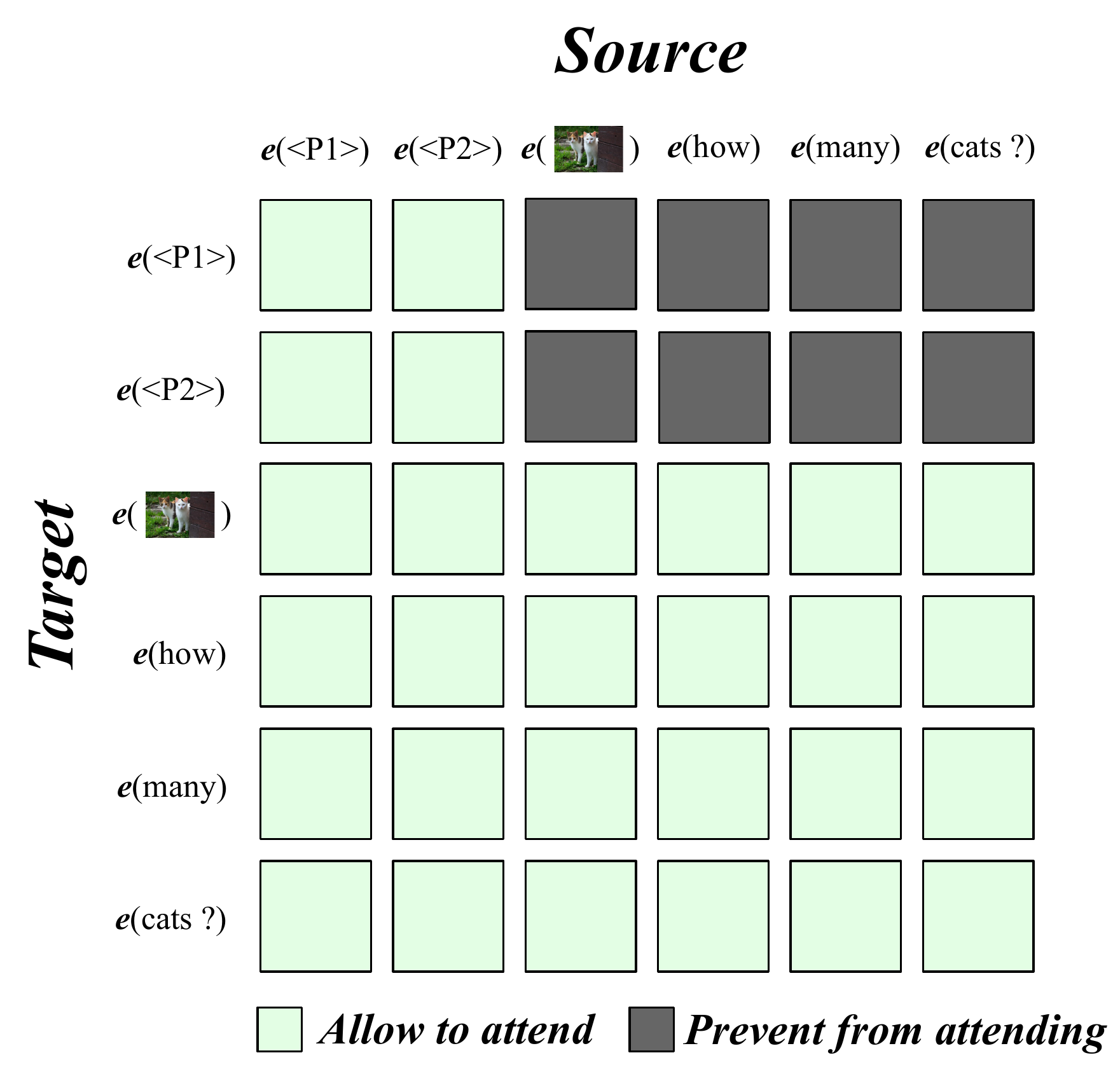}
		\caption{
			\bp attention mask in PLM encoder.
			Prompt vectors cannot attend to the
			input content, so their
			parameters solely serve to align
			the modalities.
		}
		\figlabel{mask}
	\end{figure}
	
\section{Experiments: Two Modalities}
\subsection{Setup}
\seclabel{setup}
	Our model is designed to be
	modular, maximizing the utility
	of widely used pretrained vision and language models:
	ViT \citep{dosovitskiy2021an} as our VE
	and
	BART \citep{lewis-etal-2020-bart} as our PLM.
	For both models we use the pretrained
	\emph{base} checkpoints from HuggingFace
	\citep{wolf-etal-2020-transformers}.
	We use the embedding $v$ of [CLS] as
	the image representation
	unless otherwise noted; 
	we
	use
	cross-entropy loss during training
	and use greedy search when decoding.
	
	We experiment with visual question answering
	(VQAv2; \citet{balanced_vqa_v2}), for which
 understanding
	both image and
	language
	is necessary when
	answering a question
	about an image.
	VQAv2 consists of 443,757 samples,
	categorized into three types:
	\emph{Number}, \emph{Yes/No}, and \emph{Other}.
	
	We simulate low-resource regimes by
	sampling 128 and 512 shots of training data.
	We show that \pf and \bp are less prone to overfitting
	in low-resource scenarios than baseline methods,
	in which the model tends to
        place extra emphasis on samples of the majority
        answer type \emph{Yes/No}
        but
        pays less attention to \emph{Other}.
        This is because
        the two answering words of \emph{Yes/No}
        have much
        higher
        frequency 
        in the text corpus
        than the answers
        of the open-ended questions, i.e., \emph{Other}.

	We train the models for
	two epochs on the full dataset
	and 100 epochs on the 
	sampled low-resource datasets.
	For prompting, 
	we set the prompt length $N$ to 20, and
        Appendix \secref{appendix:ablation} shows an ablation study.
        Similar to \citet{lester2021power},
        we empirically found that a large learning rate leads to
	better prompting performance. So
	we use learning rate 5e-1 for prompting;
	learning rate 5e-4 is used
	in all  other experiments.
	Batch size is  32
	and the Adam optimizer \citep{adampaper} is used.

\def\bighack{}

	\subsection{Baseline}
	
	We consider four baselines
	of fusing the modalities:

\bighack

 \textbf{Finetune}. As the baseline
		\emph{Frozen}\textsubscript{finetuned} in
		\citet{Tsimpoukelli2021MultimodalFL}, we finetune
		\emph{all parameters of VE},
                such that the
		visual embedding space is expected to be aligned
		with PLM's language embedding space.

 \textbf{Linear}. We fix VE, but train a linear layer
		to project its output, i.e., the visual embedding,
		while retaining its dimensionality. 

 \textbf{JointProj}. We concatenate
		the visual embedding $v$ to the embedding
		vector $w_i$ of each (sub)word in the sentence.
		Next, we train a linear layer to project the concatenated vectors
		to the PLM hidden dimension.
		The resulting vectors are input to the PLM encoder layers.

 \textbf{BlackImage}. To verify that the
		prompt vectors use visual information from
                VE (as opposed to simply conditioning on spurious
                features of the text, as in the above ``poodle''
                example), we train the prompt vectors with black
                images.

\bighack

	\tabref{params} shows the number of
	trained parameters of the methods.
        Finetune requires the largest number of trainable parameters,
        followed by JointProj and Linear;
	\pf and \bp are much more parameter-efficient.

\begin{table}[t]
\centering\small\renewcommand{\arraystretch}{1.2}\setlength{\tabcolsep}{4pt}
\begin{tabular}{|c|c|c|c|c|}
\hline
Finetune & Linear & JointProj & PromptFuse & BlindPrompt \\ \hline
86M      & 0.5M   & 1M        & 15K        & 15K         \\ \hline
\end{tabular}
\caption{
  Number of trainable parameters
  of different fusion methods
   in
  million (M) and thousand (K). 
}
\tablabel{params}
\end{table}

	\subsection{Results}
	\seclabel{vqaresults}
	
	\begin{table}[t]
		\centering\scriptsize
		\begin{tabular}{r|cccc}   
			\toprule       
			\textbf{Full dataset}& Other&Yes/No&Number&Overall \\
			Finetune&\textbf{20.3$\pm$0.5}&\textbf{69.3$\pm$0.3}&\textbf{29.5$\pm$0.2}&\textbf{40.1$\pm$0.3}\\
			Linear&8.5$\pm$0.6&63.9$\pm$0.2&23.3$\pm$0.3&30.1$\pm$0.3\\
			JointProj&19.2$\pm$0.4&67.7$\pm$0.2&28.9$\pm$0.4&38.9$\pm$0.1\\
			BlackImage&8.3$\pm$0.7&60.4$\pm$0.5&15.3$\pm$0.4&23.7$\pm$0.5\\
			\pf&12.2$\pm$0.6&64.9$\pm$0.4&27.1$\pm$0.2&34.1$\pm$0.4\\
			\bp&13.3$\pm$0.9&64.5$\pm$0.4&27.4$\pm$0.1&34.8$\pm$0.8\\
			
			\midrule
			\textbf{128 shots}& Other&Yes/No&Number&Overall \\
			Finetune&6.6$\pm$0.3&57.9$\pm$0.9&14.7$\pm$0.3&26.8$\pm$0.5\\
			Linear&2.3$\pm$0.1&46.4$\pm$0.7&16.2$\pm$0.4&18.2$\pm$0.4\\
			JointProj&3.9$\pm$0.5&63.3$\pm$0.1&19.4$\pm$0.6&\textbf{28.4$\pm$0.3}\\
			BlackImage&0.9$\pm$0.1&38.9$\pm$0.8&6.2$\pm$0.4&14.4$\pm$0.5\\
			\pf&4.9$\pm$0.6&\textbf{63.7$\pm$0.3}&16.9$\pm$0.2&28.3$\pm$0.6\\
			\bp&\textbf{8.0$\pm$1.1}&62.1$\pm$0.2&\textbf{19.8$\pm$0.3}&28.0$\pm$0.9\\
			
			\midrule
			\textbf{512 shots}& Other&Yes/No&Number&Overall \\
			Finetune&7.3$\pm$0.3&61.1$\pm$0.2&20.2$\pm$0.4&29.2$\pm$0.3 \\
			Linear&4.3$\pm$0.4&62.2$\pm$0.5&19.2$\pm$0.4&26.6$\pm$0.4 \\
			JointProj&3.8$\pm$0.1&63.8$\pm$0.3&\textbf{23.8$\pm$0.4}&28.7$\pm$0.3\\
			BlackImage&3.5$\pm$0.6&48.2$\pm$0.6&10.3$\pm$0.5&18.8$\pm$0.5\\
			\pf&6.3$\pm$0.5&\textbf{63.9$\pm$0.1}&21.5$\pm$0.3&29.4$\pm$0.5\\
			\bp&\textbf{8.4$\pm$0.9}&63.1$\pm$0.2&22.6$\pm$0.3&\textbf{29.7$\pm$0.6}\\
			\bottomrule
		\end{tabular}
		\caption{Results (accuracy) on VQAv2 validation set.
			We report Overall and separate
			performance of the three types of questions: Other, Yes/No, Number.}
		\tablabel{realvqa}
	\end{table}

	\tabref{realvqa} compares the performance of
	baselines and our prompting methods.
	We report mean and
	standard deviation over three runs with different
	random seeds.
	
	\pf outperforms the
	BlackImage and Linear
	baselines on all experiments, showing
	that prompting
	successfully utilizes visual information
	and fuses the two modalities.

	For 128 and 512 shots,
	\pf achieves 
	accuracy  comparable with baselines Finetune and JointProj.
	However, \emph{\pf and \bp are more parameter-efficient}
	as shown in \tabref{params}.
	Prompting methods perform worse than Finetune and JointProj
	on full data.\footnote{Finetune (40.1) performs worse than
          Frozen\textsubscript{VQA} (48.4). We hypothesize this is because
          \fz uses a much larger PLM (7 billion) than ours (139 million).
        }
        We conjecture that this is due to
	having much fewer
	parameters, i.e., 15K, which is even smaller than 
	the training set size   443,757.
	Thus we argue that {\pf} better suits
	low-resource scenarios.

	In
	low-resource
	experiments,
	\pf and \bp
	achieve higher
	accuracy on \emph{Other} and \emph{Number};
	the performance drops
	on \emph{Yes/No} compared with Finetune and JointProj.
	This also happens between \pf and \bp. For example, on 128 shots,
        we find that
	\bp outperforms \pf with
        3\% on \emph{Number} and 3\% on \emph{Other}.
	The results indicate
        that our prompting methods,
	especially \bp, can better utilize the
        generalization capability
	of PLM to handle open-ended questions and
        are less prone to
        falling
        into \emph{Yes/No} samples.

\begin{table*}[t]
	\centering\scriptsize
	\begin{tabular}{c|ccc}
		\toprule
		& NoPrompt&\pf&\bp \\ \midrule
		\begin{minipage}[b]{0.4\columnwidth}
			\centering
			\raisebox{-.5\height}{\includegraphics[width=\linewidth]{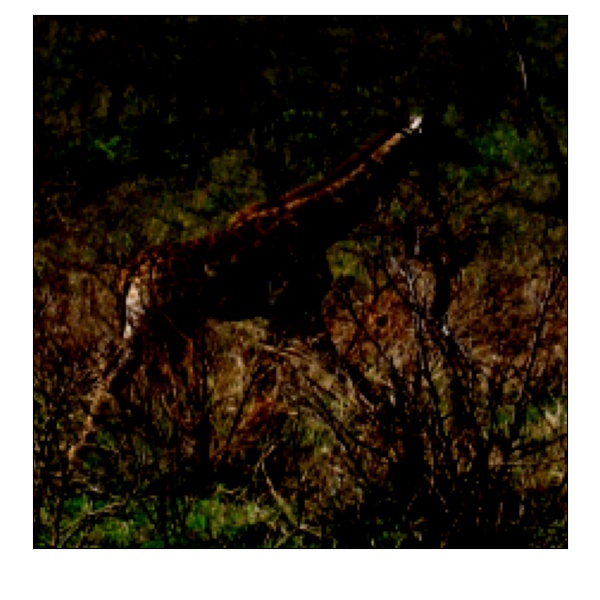}}
		\end{minipage}
		&
		\begin{minipage}[b]{0.4\columnwidth}
			\centering
			\raisebox{-.5\height}{\includegraphics[width=\linewidth]{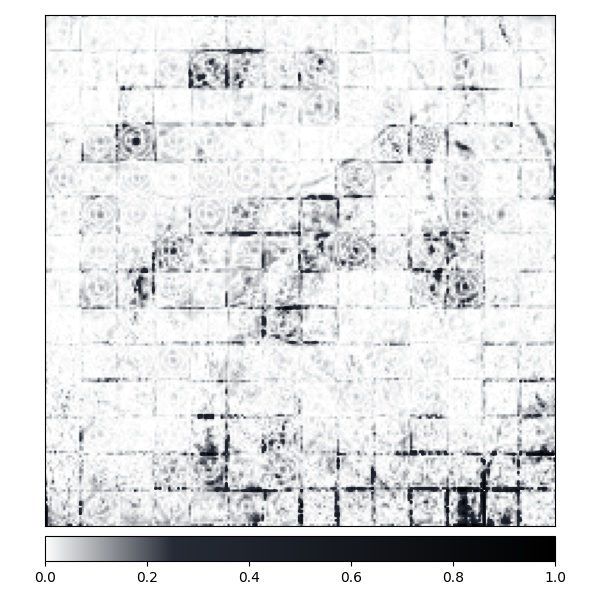}}
		\end{minipage}
		&
		\begin{minipage}[b]{0.4\columnwidth}
			\centering
			\raisebox{-.5\height}{\includegraphics[width=\linewidth]{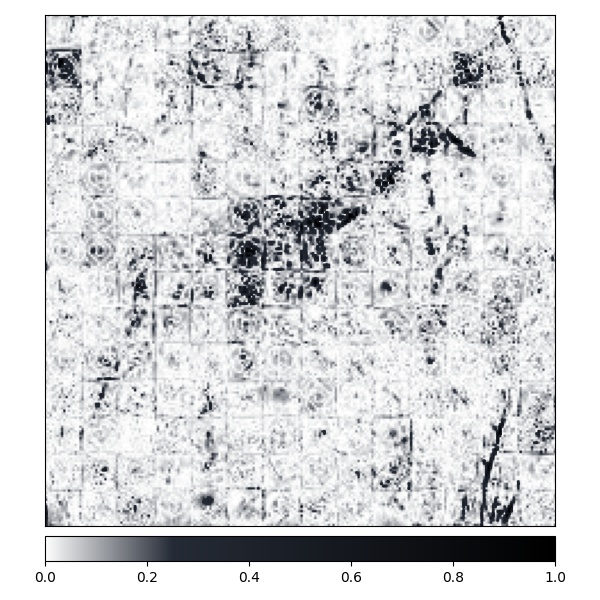}}
		\end{minipage}
		&
		\begin{minipage}[b]{0.4\columnwidth}
			\centering
			\raisebox{-.5\height}{\includegraphics[width=\linewidth]{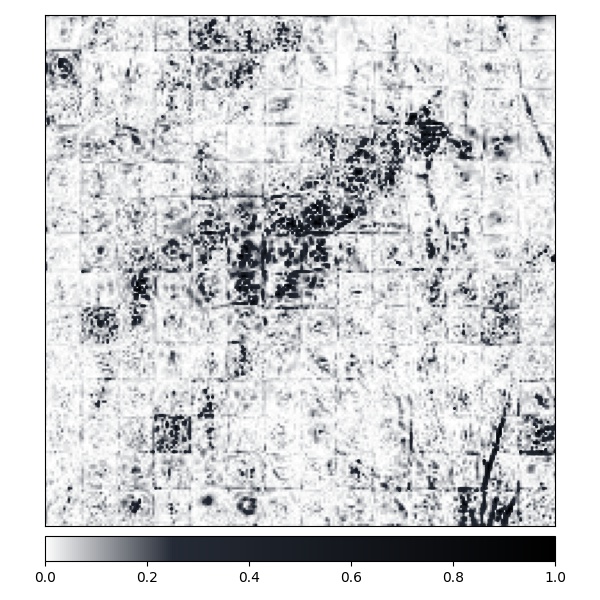}}
		\end{minipage}
		\\ \midrule
		Question
		&\hlc[cyan!326]{Do} \hlc[cyan!31]{you} \hlc[cyan!2]{see} \hlc[cyan!9]{a} \hlc[pink!1]{gir}\hlc[cyan!14]{affe} \hlc[cyan!6]{in} \hlc[cyan!23]{the} \hlc[pink!5]{picture}\hlc[cyan!222]{?}
		&\hlc[cyan!5]{Do} \hlc[pink!90]{you} \hlc[pink!50]{see} \hlc[cyan!50]{a} \hlc[cyan!57]{gir}\hlc[pink!6]{affe} \hlc[cyan!14]{in} \hlc[cyan!9]{the} \hlc[pink!190]{picture}\hlc[pink!241]{?}
		&\hlc[cyan!20]{Do} \hlc[pink!10]{you} \hlc[cyan!10]{see} \hlc[pink!10]{a} \hlc[pink!20]{gir}\hlc[pink!10]{affe} \hlc[pink!10]{in} \hlc[pink!10]{the} \hlc[pink!10]{picture}\hlc[pink!20]{?}
		\\ \midrule
		Prediction
		&\textless /s\textgreater
		&Yes
		&Yes
		\\
		\bottomrule
	\end{tabular}
	\caption{Attribution score magnitude heat map for image and text inputs.
		Black/white image pixels indicate positive/negative influence on
		predicting ``Yes'',
		and the same goes for red/blue tokens.
		Integrated gradients are calculated only on the first prediction after decoder input ``\textless /s\textgreater \textless s\textgreater'' in an auto-regressive manner.}
	\tablabel{ig}
\end{table*}

\subsection{Qualitative Example}
To understand how prompting helps in fusing different modalities,
we compare \pf and \bp to a \textbf{NoPrompt} baseline.
NoPrompt directly concatenates the
visual outputs from VE to the text
input of the PLM without any training.

Concretely,
we apply the \textbf{Integrated Gradients} method \citep{sundararajan2017axiomatic},
which measures the attribution of features to the neural network outputs.
Traditional approaches define feature importance
by the gradient of model outputs to input features.
Integrated gradients extend this measure as the
path integral of the gradient from a baseline -- reflecting the absence of signal -- to
the actual input. In practice, we use the Captum
package \citep{kokhlikyan2020captum} in our implementation.

\tabref{ig} illustrates a qualitative example when applying
NoPrompt, \pf, and \bp on VQAv2.
For NoPrompt, because no training is involved,
visual embeddings from VE confuse the PLM, leading to a wrong
prediction (``\textless /s\textgreater''). The system is not able to correctly
understand the image and question.
In contrast, \pf and \bp guide the PLM to pay
attention to the
image and identify the regions of ``giraffe''
and then correctly respond ``Yes''.

Interestingly, the attribution scores of the question from \bp
are small, compared to \pf. We conjecture the reason is
that, understanding the question -- which has a straightforward
syntactic/semantic structures -- is relatively simple for the
PLM because it has been pretrained on a large volume of
text. \bp thus enforces that the multimodal system  focus more
on the visual embeddings (i.e., the encoded image), which
is a new source of information for answering the question.

\section{Experiments: Three Modalities}
\seclabel{threemod}
Disentangling functionality
of the modality data encoder, e.g., VE,
makes
\pf and \bp
more modular than \fz.
Applying
our methods to
tasks involving more
than two modalities
is straightforward.
In contrast, \fz incurs
the high cost
of pretraining encoders
for new modalities.
We experiment
on the sarcasm detection dataset MUStARD \citep{mustard} with
video, audio, and text data.\footnote{
  To highlight modularity, we
  utilize pretrained encoders rather than
  the data preprocessing pipelines in \citet{mustard}.
  For example, we use 
  pretrained wav2vec2 \citep{baevski2020wav2vec}
  rather than Mel-Frequency Cepstral Coefficients \citep{MFCC}
  when processing audio data.
}

\textbf{Setup}.
To process video, we first use OpenFace
\citep{openface} to sample important
frames containing human faces.
Next, ViT is leveraged to
extract visual representations
from each frame.
We then average 
visual representations of
all frames to represent the video.
To process audio,
we use librosa \citep{mcfee2015librosa}
to remove background noise
and convert audio to
waveform with a sampling rate of 16,000 Hz.
We then use
pretrained
wav2vec2 \citep{baevski2020wav2vec}
to encode the waveform
and apply the same
averaging strategy as for video.
BART is used as our PLM.
We use a verbalizer of \emph{True/False}
in this experiment.

We adopt the speaker-dependent setup in MUStARD:
334 training and
356 testing samples. 
We compare \pf, \bp, and Finetune
for 8, 32, and 64 shots.
Note that Finetune uses
180M trainable parameters 
in the vision and audio encoders.
We also conduct an experiment
training on the full dataset for 5 epochs.
The remaining setup is the same as \secref{setup}.

	\begin{table}[t]
		\centering\scriptsize
		\begin{tabular}{r|ccc}   
			\toprule       
			\textbf{Full dataset}& Precision&Recall&F-Score \\
			Finetune&65.6$\pm$0.2 &73.9$\pm$2.7 &\textbf{68.4}$\pm$\textbf{0.5} \\
			\pf&64.2$\pm$0.4 &72.1$\pm$3.6 &66.2$\pm$0.7\\
			\bp&63.8$\pm$0.5 &71.9$\pm$3.1 &66.5$\pm$0.8\\
			
			\midrule
			\textbf{8 shots}& Precision&Recall&F-Score \\
			Finetune&42.8$\pm$4.3 &69.5$\pm$9.9 &52.7$\pm$5.5 \\
			\pf&41.1$\pm$4.8 &71.0$\pm$13.1 &53.1$\pm$5.8\\
			\bp&44.2$\pm$4.5& 71.8$\pm$12.8& \textbf{54.0}$\pm$\textbf{6.1}\\
			
			\midrule
			\textbf{32 shots}& Precision&Recall&F-Score \\
			Finetune&53.9$\pm$4.1 &70.6$\pm$9.1 &\textbf{59.1}$\pm$\textbf{5.2} \\
			\pf&53.8$\pm$4.7 &71.1$\pm$10.8 &58.5$\pm$5.4\\
			\bp&54.6$\pm$4.1 &69.7$\pm$10.3 &58.7$\pm$5.5\\
			
			\midrule
			\textbf{64 shots}& Precision&Recall&F-Score \\
			Finetune&59.5$\pm$2.3 &70.4$\pm$7.7 &61.4$\pm$2.8 \\
			\pf&59.2$\pm$2.7 &70.2$\pm$7.4 &\textbf{62.0}$\pm$\textbf{3.3}\\
			\bp&60.1$\pm$2.4 &70.9$\pm$7.8 &61.7$\pm$3.1\\
			\bottomrule
		\end{tabular}
		\caption{ Results on MUStARD test set.
                }
		\tablabel{mustard}
	\end{table}

\textbf{Results}.
 \tabref{mustard} reports performance over ten runs.
\pf and \bp outperform Finetune
in 8- and 64-shot experiments.
Prompting methods perform
comparably
to Finetune in other
experiments,
\emph{while they
are clearly more parameter-efficient.}
Overall, the
three-modality experiment
provides observations in line with \secref{vqaresults}.
More importantly,
it highlights two
strengths of prompting:
High modularity and
parameter-efficiency.

\section{Conclusion}
We propose \pf and \bp as methods for aligning different
modalities in a modular and parameter-efficient manner.  We
show that prompting, which requires only a few trainable parameters,
performs comparably to several multimodal fusion methods
in low-resource scenarios.
The high modularity property of prompting supports -- by
avoiding the need to finetune large pretrained models --
flexible addition of modalities at
low cost.


\section*{Acknowledgements}
This work was supported by the European Research Council (\# 740516). We thank the
anonymous reviewers for valuable comments.
\bibliography{anthology,custom}
\bibliographystyle{acl_natbib}

\clearpage
\appendix
\section{Ablation Analysis}
\seclabel{appendix:ablation}
As an ablation analysis, we test
variants of \pf and \bp
with full data on VQAv2 dataset.
All experiment setup follows \secref{setup}.

\textbf{Prompt length}.
\pf and \bp have an extremely limited number of trainable parameters,
making it challenging
to achieve performance as finetuning
in high-resource scenarios.
Intuitively, we would like to inject more prompt vectors
to increase the number of trainable parameters.
\tabref{pormptlen} shows that both \pf and \bp obtain best
accuracy when the prompt length is set to 60.
Using a particularly large length (e.g., 100)
harms performance.
This is in line with \citet{lester2021power}:
They find that too much prompt information may bring negative effects.
Since more prompt vectors also consume more training time, we use 20 in our experiments.

\begin{table}[h]
        \centering\scriptsize
        \begin{tabular}{r|ccccccc}
                \toprule
                &5&10&20&40&60&80&100\\
                \midrule
                \pf & 28.5&30.4&34.1&35.3&\textbf{35.8}&34.2&30.3\\
                \bp &27.1&30.7&34.8&35.5&\textbf{35.6}&34.4&30.9\\
                \bottomrule
        \end{tabular}
        \caption{Overall accuracy on VQAv2 validation set with
        prompt length ranging from 5 to 100.
        We report mean performance over three random seeds.}
        \tablabel{pormptlen}
\end{table}

\textbf{Prompt position}.
In this work we inject prompt vectors at the beginning
of input fed to PLM (see \figref{model}),
here we test two alternative positions for injection:
(i) middle, i.e., inserting between vision and (sub)word embeddings;
(ii) end of the question.
Results in \tabref{pormppos} show that these positions
yield similar performance,
indicating that our approach is not largely affected by prompt positions.

\textbf{Prompt encoder}.
Another approach to increase trainable parameters
is to use an extra module to encode prompt vectors.
We test two neural network modules:
(i) a linear layer; (ii) an LSTM \citep{lstmpaper}.
Both modules have the same hidden dimension as the PLM.
However, these variants only bring small improvements,
as presented in \tabref{pormppos}.
Future work may explore more advanced methods
of scaling up the number of parameters.

\textbf{Visual embedding}.
In addition to utilizing the [CLS] embedding,
there are two alternative 
ViT outputs can be used as the
visual embeddings:
(i) the entire embedded sequence;
(ii) the embedding averaged over the sequence.
\tabref{pormppos} shows that these
approaches achieve comparable results.
To save computational resources, we use [CLS] for images in VQAv2.
For video frames and speech signals in MUStARD,
we use average due to large sequence lengths.

\begin{table}[t]
        \centering\scriptsize
        \begin{tabular}{c|c|cc}
                \toprule
                &&\pf&\bp\\
                \midrule
                &Baseline&34.1$\pm$0.4&34.8$\pm$0.8\\
                \midrule
                Prompt&Middle &33.7$\pm$0.4&34.9$\pm$0.7\\
                Position&End &34.3$\pm$0.5&34.5$\pm$0.6\\
                \midrule
                Prompt &Linear &34.7$\pm$0.5&35.0$\pm$0.6\\
                Encoder&LSTM &34.9$\pm$0.4&35.1$\pm$0.4\\
                \midrule
                Visual &Seq & 34.6$\pm$0.6& 34.7$\pm$0.5\\
                Embedding&Avg &33.9$\pm$0.5&34.9$\pm$0.4\\
                \midrule
        \end{tabular}
        \caption{Results on VQAv2 validation set
                with variants of prompt position,
                encoder, and visual embedding.}
        \tablabel{pormppos}
\end{table}

\section{Modularity}
This section further demonstrates
the modularity and flexibility
of \pf and \bp.
Besides the ability of utilizing  encoders of more than
two modalities as shown in \secref{threemod},
the modular design allows \pf and \bp to use PLMs other than BART.
Concretely, we compare BERT/T5 to BART,
by full data training on VQAv2 as \secref{setup}.
BERT is a masked language model,
thus we train and evaluate only on \emph{Number} and \emph{Yes/No} samples,
by filling the mask in pattern ``Question: \emph{input question} Answer: [MASK]''.

As reported in \tabref{otherPLMs}, BERT performs well on
\emph{Number} and \emph{Yes/No} compared to BART,
indicating that \pf/\bp can
also be applied to encoder-only architecture.
Also, T5 outperforms BART, especially on \emph{Other},
further indicating that \pf/\bp are compatible with new
PLMs, which give increasingly better task performance.

\begin{table}[t]
        \centering\scriptsize
        \begin{tabular}{r|cccc}
                \toprule
                \textbf{BART}& Other&Yes/No&Number&Overall \\
                \pf&12.2$\pm$0.6&64.9$\pm$0.4&27.1$\pm$0.2&34.1$\pm$0.4\\
                \bp&13.3$\pm$0.9&64.5$\pm$0.4&27.4$\pm$0.1&34.8$\pm$0.8\\

                \midrule
                \textbf{BERT}& Other&Yes/No&Number&Overall \\
                \pf&-&67.5$\pm$0.3&28.4$\pm$0.2&-\\
                \bp&-&67.8$\pm$0.4&28.6$\pm$0.2&-\\

                \midrule
                \textbf{T5}& Other&Yes/No&Number&Overall \\
                \pf&15.8$\pm$0.7&65.4$\pm$0.2&27.3$\pm$0.3&36.5$\pm$0.4\\
                \bp&16.2$\pm$0.8&65.2$\pm$0.3&27.4$\pm$0.2&36.6$\pm$0.6\\
                \bottomrule
        \end{tabular}
        \caption{Results with BERT and T5 on VQAv2 validation set.}
        \tablabel{otherPLMs}
\end{table}

\section{Experiment Setup}
\begin{table*}[t]
	\centering\scriptsize
	\begin{tabular}{r|rrrrrrrrr}   
		\toprule       
		Dataset&Modalities
		&\# Train&\# Test
		&Runs
		&Batch Size
		&Epochs 
		&Prompt Length
		&LR (Prompt)
		&LR (Other) \\
		\midrule
		VQAv2&Image, Text&443,757&214,354&3&32&2&20&5e-1&5e-4\\
		low-resource&Image, Text&128/512&214,354&3&32&100&20&5e-1&5e-4\\
		\midrule
		MUStARD&Video, Audio, Text&334&356&10&8&5&20&5e-1&5e-4\\
		low-resource&Video, Audio, Text&8/32/64&356&10&8&50&20&5e-1&5e-4\\
		\bottomrule
	\end{tabular}
	\caption{Dataset statistics and hyperparameters
          used in the experiments.
	}
	\tablabel{summary}
\end{table*}
\tabref{summary} shows the setup used in all of our experiments.
We use 8 GEFORCE GTX 1080Ti GPUs and
gradient accumulation is applied during training.

\end{document}